\pdfoutput=1 
\documentclass{article}

\usepackage{ifpdf}
\usepackage{spconf}
\usepackage{graphicx}
\usepackage[table,xcdraw]{xcolor}
\usepackage[square,numbers]{natbib}
\setcitestyle{square}
\usepackage[utf8]{inputenc} % allow utf-8 input
\usepackage[T1]{fontenc}    % use 8-bit T1 fonts
\usepackage{hyperref}       % hyperlinks
\usepackage{url}            % simple URL typesetting
\usepackage{booktabs}       % professional-quality tables
\usepackage{amsfonts}       % blackboard math symbols
\usepackage{nicefrac}       % compact symbols for 1/2, etc.
\usepackage{microtype}      % microtypography
\usepackage{xcolor}         % colors
\usepackage{comment}
\usepackage{tabularx}       % tables
\usepackage{subcaption}
\usepackage{afterpage}
\usepackage{pdflscape}
\usepackage{float} 
 
\usepackage{booktabs} % for multipage tables
\usepackage{multirow} 
\usepackage{adjustbox}
\usepackage{longtable}
\usepackage{lscape}
\usepackage{placeins}
\usepackage{wrapfig,lipsum,booktabs}
\usepackage[bottom]{footmisc}

\renewcommand{\thefootnote}{\fnsymbol{footnote}}

\title{Pseudo-Inverted Bottleneck Convolution for DARTS Search Space}
\name{\begin{tabular}{c}Arash Ahmadian$^\textit{1,*}$\, Louis S.P. Liu$^\textit{1,*}$, Yue Fei$^\textit{1}$, Konstantinos N. Plataniotis$^\textit{1}$, Mahdi S. Hosseini$^\textit{2}$
\end{tabular}\vspace{-4mm}}
\address{%
\small $^\textit{1}$The Edward S. Rogers Sr. Department of Electrical \& Computer Engineering, University of Toronto, Canada  \vspace{-1mm} \\
\small $^\textit{2}$Computer Science and Software Engineering (CSSE), Concordia University, Canada
\vspace{-3mm}
}

\begin{document}\sloppy
\maketitle
\renewcommand{\thefootnote}{\fnsymbol{footnote}}
\footnotetext[1]{Equal Contribution}

\begin{abstract}
Differentiable Architecture Search (DARTS) has attracted considerable attention as a gradient-based neural architecture search method. Since the introduction of DARTS, there has been little work done on adapting the action space based on state-of-art architecture design principles for CNNs. In this work, we aim to address this gap by incrementally augmenting the DARTS search space with micro-design changes inspired by ConvNeXt and studying the trade-off between accuracy, evaluation layer count, and computational cost. We introduce the Pseudo-Inverted Bottleneck Conv (PIBConv) block intending to reduce the computational footprint of the inverted bottleneck block proposed in ConvNeXt. Our proposed architecture is much less sensitive to evaluation layer count and outperforms a DARTS network with similar size significantly, at layer counts as small as $2$. Furthermore, with less layers, not only does it achieve higher accuracy with lower computational footprint (measured in GMACs) and parameter count, GradCAM comparisons show that our network can better detect distinctive features of target objects compared to DARTS. Code is available from \url{https://github.com/mahdihosseini/PIBConv}.
\end{abstract}

\section{Introduction}
Since the introduction of Vision Transformers (ViTs) by \textit{Dosovitskiy et al.} \cite{vit}, a new class of research has emerged, pushing the boundaries of Transformer-based architectures on a variety of computer vision tasks \cite{distillation, swin, pyramidVIT, tokens_to_tokens}. These advances make it seem inevitable that ViTs would overtake conventional Convolutional Neural Networks (CNNs). Recently, \textit{Liu et al.}’s ConvNeXt \cite{convnext} has sparked a resurgence in further exploring the architectural designs of CNNs in image recognition. Specifically, they argued that by adapting components from Transformers into the standard ResNet backbone \cite{ResNet}, the trained models can match or outperform state-of-the-art ViTs in image classification, objection detection, and segmentation. If CNNs can still be improved by design elements that were previously overlooked, this begs the question: \textit{Can we apply the same Transformer principles to a Neural Architecture Search (NAS) framework to improve its performance?}

NAS has historically seen immense success on large-scale image classification prior to ViTs \cite{EfficientNet, EfficientNetV2, SearchedMobileNetV3} as it alleviates the task of manually designing for the optimal neural network architecture. Early works of NAS employed Reinforcement Learning \cite{ZophRL}, Evolutionary Search \cite{Hierar-repres}, and Bayesian Optimization \cite{FNAS} while more recent works have shifted to the One-Shot NAS paradigm \cite{SPOS}. One popular branch stream of NAS is Differentiable Architecture Search (DARTS) \cite{DartsLiu}. It relaxes the search space from discrete to continuous by attributing weights to each operation in the set using a \textit{Softmax} function and choosing the best candidate. In DARTS, a \textit{$n$-layer} network is constructed by replicating a \textit{normal cell}, $n$ times and adding \textit{reduction cells} at the $1/3$ and $2/3$  of the total depth with. We refer the reader to \cite{DartsLiu} for more details. 

Several works investigate improving the NAS operation space using methods such as increasing the granularity of operations by breaking down search units across input channels \cite{AtomNAS}, grouping similar operations to combat the effects of multi-collinearity \cite{StacNAS}, creating more expressive operations by replacing the DFT matrices in convolution's diagonalization with K-matrices \cite{xd_operations}, and reducing the operation set \cite{nasbench201}. Here, we investigate optimizations to the search space through a different set of lens by drawing inspiration from ConvNeXt.

We start with the second-order DARTSV2 cell (vanilla) structure and incrementally augment the search operations by adapting design elements from ConvNeXt. For each stage, we conduct search and evaluation phases on CIFAR-10 \cite{cifar10} using the same training setup and hyper-parameters as DARTS \cite{DartsLiu}. In our experiments, we encountered a large increase in parameter count when directly adopting the ConvNeXt convolution block with hindering performances. To combat this, we propose Pseudo-Inverted Bottleneck Convolution (PIBConv) structure to incorporate an inverted bottleneck while minimizing model size. Our proposed architecture is much less sensitive to evaluation layer count and achieves better test error than the original DARTSV2 with comparable parameter count and computations. We further demonstrate its effectiveness by performing a GradCAM \cite{gradcam} analysis, showing that it is able to capture prominent image features at 10 layers vs. a 20-layer DARTSV2. Our contributions are:

[\textbf{C1.}] We present an incremental experiment procedure to evaluate how design components from ConvNeXt impact the performance of DARTS by redesigning its search space.

[\textbf{C2.}] We introduce PIBConv block to implement an inverted bottleneck structure while minimizing model footprint and computations. This outperforms vanilla DARTSV2 with lower layer count, parameter count, and GMACs.

\section{Methodology}
Our approach to modernizing the DARTS operation set involves incrementally making micro-changes to the design of the separable conv block used within DARTS. However, not all changes proposed in ConvNeXt can be transferred to DARTS. (1) \textit{Changing the stage compute ratio} to match that of the Swin Transformer \cite{swin} is not applicable as it would require major restructuring of the DARTS framework (\textit{i.e.} changing the placement of reduction cells) which is beyond our scope of updating the operation set. (2) \textit{Modifying the stem cell} to mimic the ``patchify'' operation in Swin is not applicable since a $4\times$ downsampling is too aggressive for the $32\times 32$ images in CIFAR-10. With every change, we search for a cell structure (or \textit{genotype}), under hyper-parameter settings described in Section 4 and evaluate on different layer counts (\ref{tab:comparison}).  We compare the highest achieved accuracies and corresponding GMACs. Below we present this exploration step by step. Note that incremental
changes are accumulated from step-to-step unless otherwise
stated explicitly. 

\begin{figure}[htp]
\label{figure:methodology_acc}
    \graphicspath{ {./images/}}
        \includegraphics[scale=0.11]{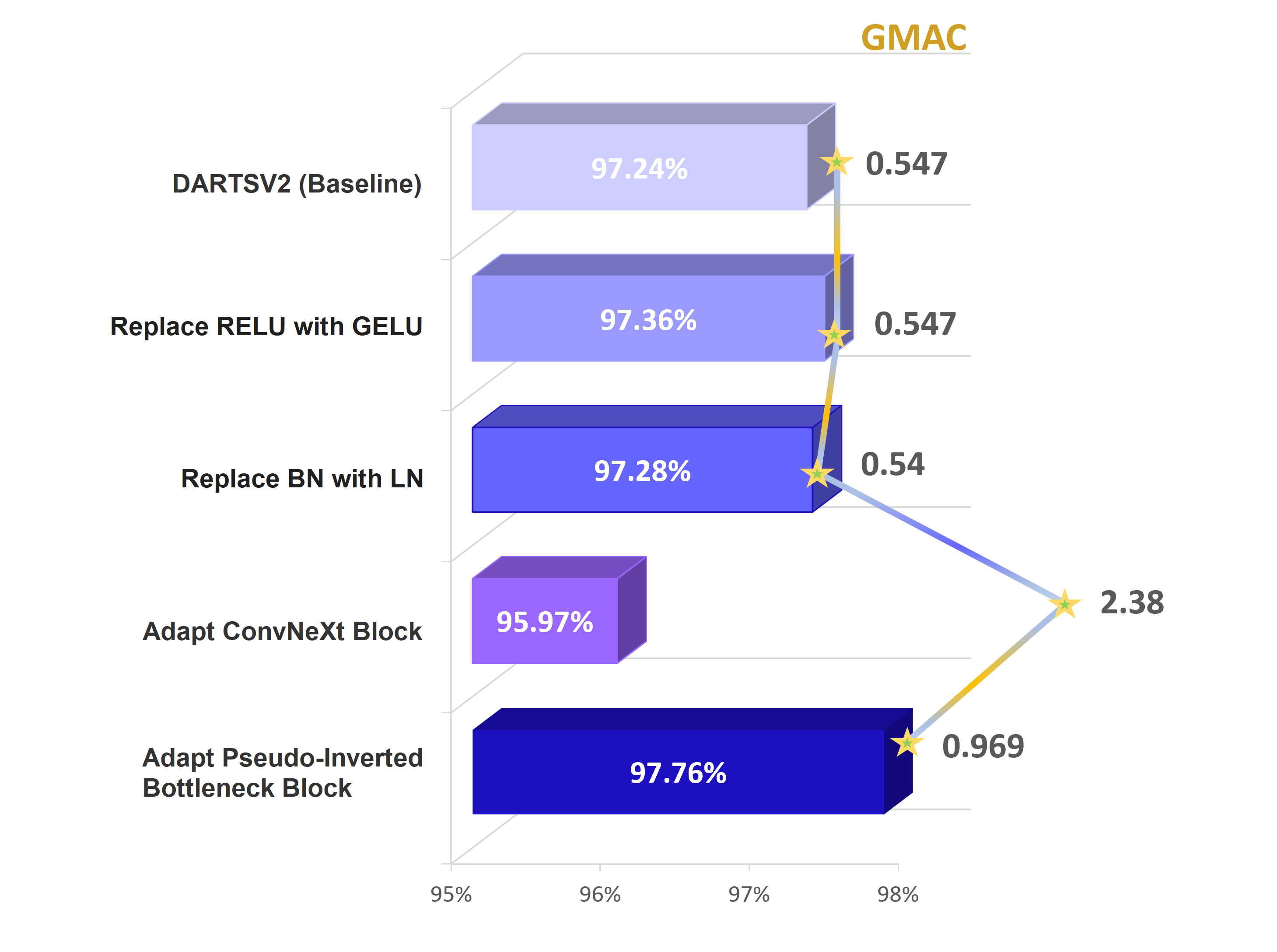}
    \caption{Roadmap of the incremental augmentations described in Section 3, along with their corresponding accuracies and methodologies.}
\end{figure}

\textbf{Replacing ReLU with GELU} We replace the widely used ReLu \cite{RELU} activation with GELU \cite{GELU} which provides an approximation of the former with the key distinction that a small portion of negative signals are let through to the next layer. \textit{This boosts the accuracy by $0.12\%$ and from now on we use GELU instead of ReLU.}
 
\textbf{Replacing BatchNorm with LayerNorm} There have been multiple attempts to develop an alternative to normalization however it remains a key ingredient in modern NN design \cite{NIPS2016_ed265bc9}. In ConvNeXt, replacing BN with LN slightly improves the accuracy of the network. We replace BatchNorm \cite{batchnorm} with LayerNorm \cite{layernorm} in our separable convolution operation. Initially, this results in minor degradation in accuracy. We also experiment with retaining LN and adding the various micro-changes proposed in this section. We did not achieve a performance boost from LN in any setting. \textit{We will use BN instead of LN.}
	
\definecolor{LightCyan}{rgb}{0.67, 0.89, 0.996}

\begin{figure}[htp]
    \graphicspath{ {./images/}}
    \centering
    \includegraphics[scale=0.068]{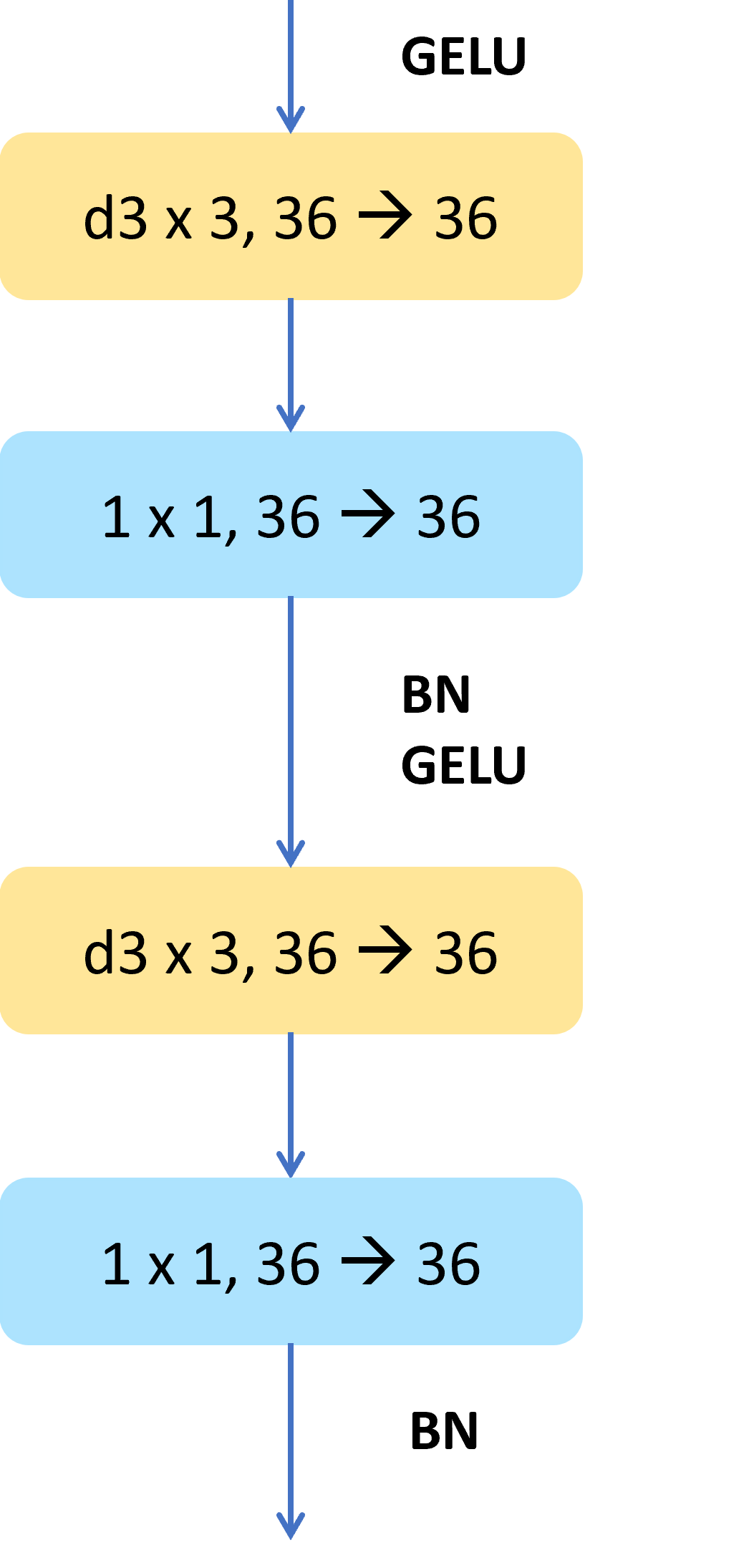}\label{fig:darts_norm}
    \includegraphics[scale=0.23]{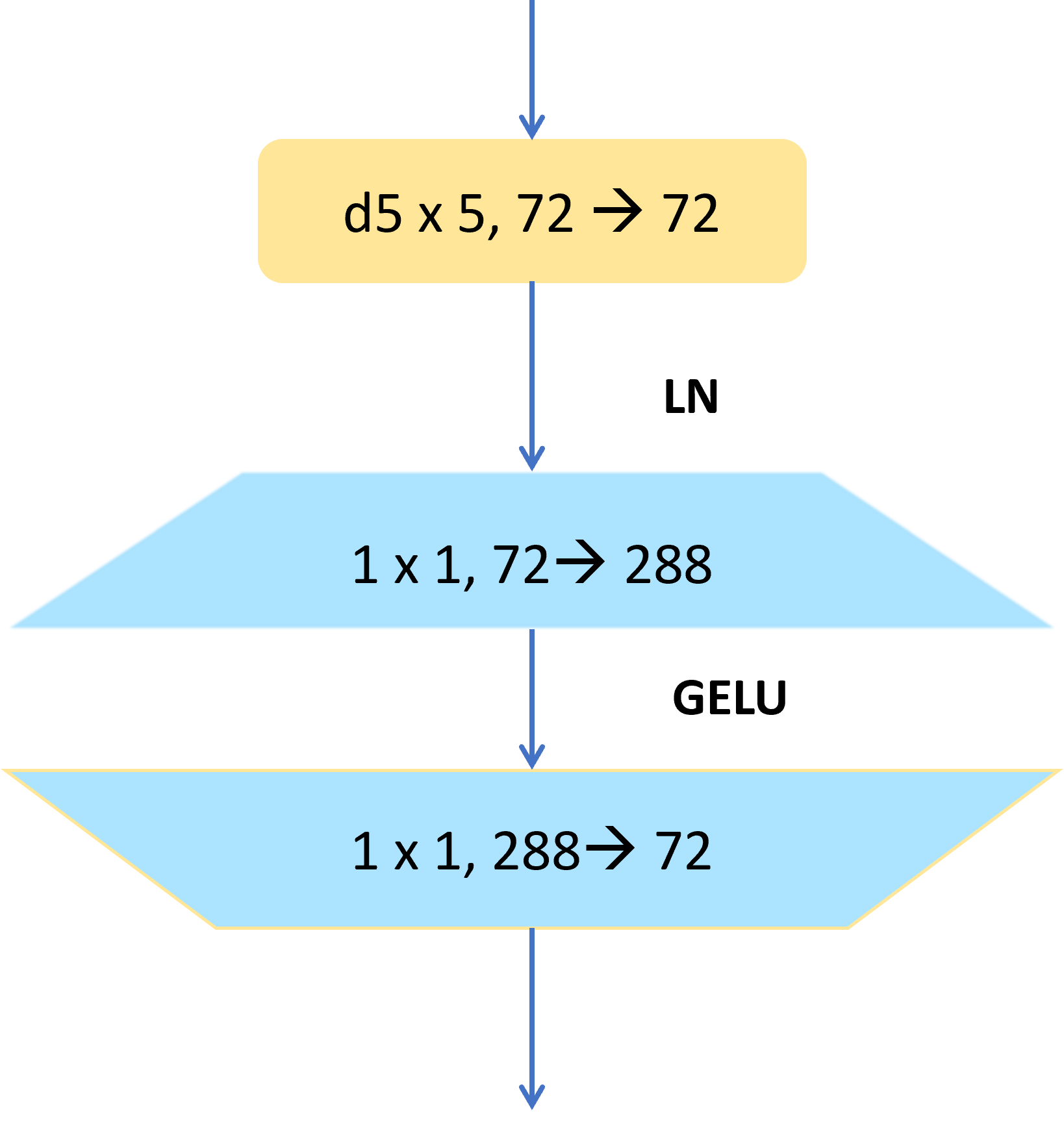}\label{fig:convnext_norm}
    \includegraphics[scale=0.234]{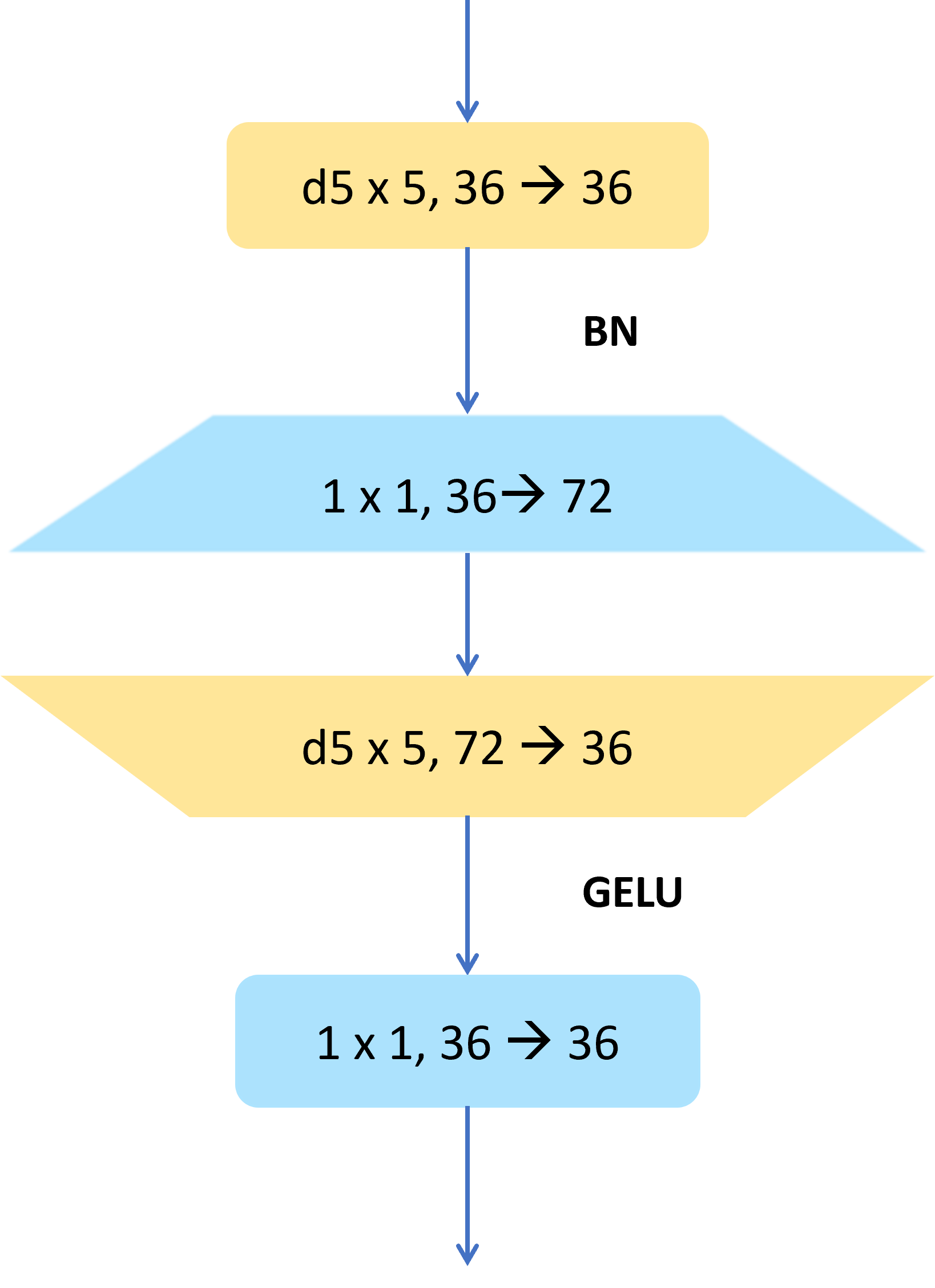}\label{fig:newconv_norm}
    \caption{Convolution Blocks : \textbf{(a)} DARTS Separable Convolution Block; \textbf{(b)} Inverted Bottleneck ConvNeXt Convolution Block ($C_{inv} = C\times 4$); \textbf{(c)} Pseudo-Inverted Bottleneck Cell ($C_{inv} = C \times 2$)}
    \label{fig:convblk}
\end{figure}

\textbf{Adapting the ConvNeXt Block} Vanilla DARTS uses depthwise separable convolution as popularized by Xception \cite{xception}. The stacked topology used in DARTS is depicted in Fig. \ref{fig:convblk}a. However, the inverted bottleneck popularized by MobileNetV2 \cite{mobilenetv2} has made its way to multiple modern networks \cite{EfficientNet, MnasNet} and thus warrants exploration in the DARTS framework. We implement the ConvNeXt block structure in Fig. \ref{fig:convblk}b (refer to \cite{convnext} for further details on the reasoning behind the architectural design choices). It consists of three key changes: (1) Reducing the number of activation and normalization functions, (2) Adapting to an inverted bottleneck structure, and (3) Moving up the depthwise separable conv layer to facilitate training with large kernel sizes. However, directly adapting the ConvNeXt block significantly increases the number of parameters and GMACs while sharply decreasing accuracy.

To manage the number of learnable parameters,  we introduce PIBConv (Pseudo-Inverted Bottleneck Conv block) as depicted in Fig. \ref{fig:convblk}. We add a depthwise convolution after the intermediate pointwise conv layer which reduces the number of channels. We keep the positions of the activation and normalization the same relative to the next layer based on the ConvNeXt block. This structure also inhibits the stacked architecture which has been shown to increase accuracy by $1-2\%$ when introduced to separable convolution-based operations in the DARTS framework \cite{DartsLiu} (which the vanilla inverted bottleneck does not have), as well as an inverted bottleneck structure. 

We compare the number of weights per block to estimate the parameter size and computational complexity of both networks. Define $C$ to be the input and output channel size, $C_{inv}$ to be the inverted bottleneck channel size, and $K$ to be the kernel size of the depthwise convolution. Similarly, define $F = C_{inv}/C$ to be the inverted bottleneck ratio for the first pointwise convolution. The total number of weights between the ConvNeXt block (1) and our PIBConv block (2) are compared below:
\begin{equation} \label{darts_invBN_conv_size}
2FC^2+K^2C
\end{equation}
\begin{equation} \label{eq:bottleneck_eq}
(F+1)C^2+2K^2C
\end{equation}

In practice, the dominant variable in both equations is the channel size $C$, which is initialized to $16$ and doubled at each reduction cell. Additionally, the conv operation dominates both DARTSV2 and our searched genotypes. Thus, comparing the coefficients of the quadratic term $C^2$ provides an estimate for the difference in parameter size and computational complexity of these networks. Our PIBConv block has approximately $0.63$ times the number of weights as the ConvNeXt block. We further choose $F=2$ in the final block topology after experimentation with various values in $\{1.5, 4.5\}$ since it achieved the best accuracy-GMAC trade-off. \textit{The use of the Pseudo-Inverted Bottleneck block boosts the accuracy by $0.4\%$}.

\section{Experiments}
\textbf{Experimental Setup} We present our hyperparameter settings and experimental setup next. Following the DARTS framework, we search with an initial channel size of $16$, $4$ nodes, $8$ layers, $50$ epochs, and a batch size of $64$. We use the SGD optimizer coupled with a cosine-annealing learning rate scheduler (no restarts) \cite{cosineAnnealing}, $0.0025$ initial learning rate, $3e-4$ weight decay, and $0.9$ momentum. As for the evaluation phase, we train for $600$ epochs with a batch size of $96$, cutout augmentation \cite{cutout}, path dropout with probability $0.2$ and auxiliary towers with $0.4$ weight. Other hyper-parameter settings remain the same as the search phase. Both our search and evaluation phases are performed on CIFAR-10.

\begin{figure*}[htp]
    \graphicspath{ {./images/}}
    \scriptsize
    \centering
    \includegraphics[height=0.25\linewidth]{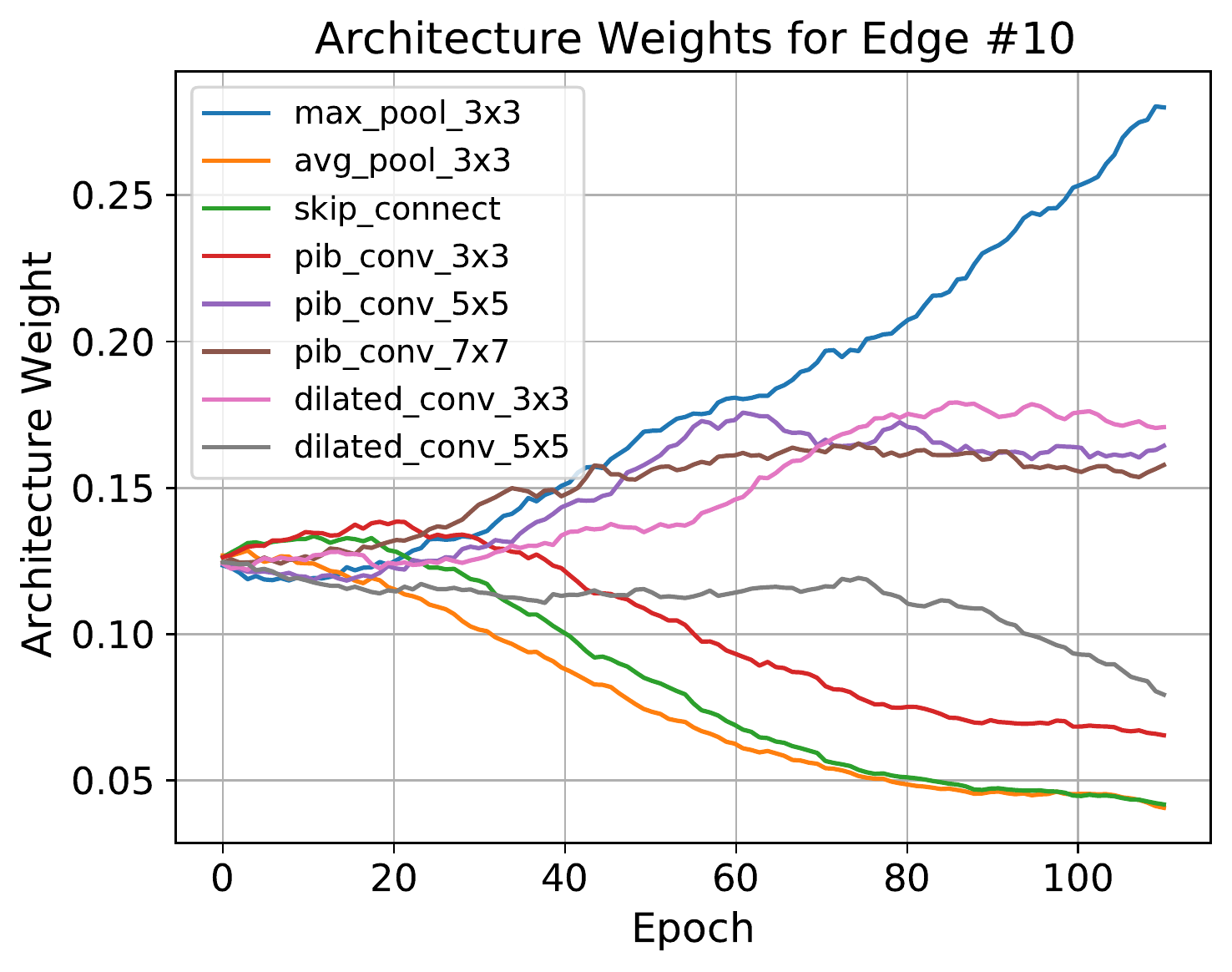}
    \hspace{0.2cm}
    \includegraphics[height=0.25\linewidth]{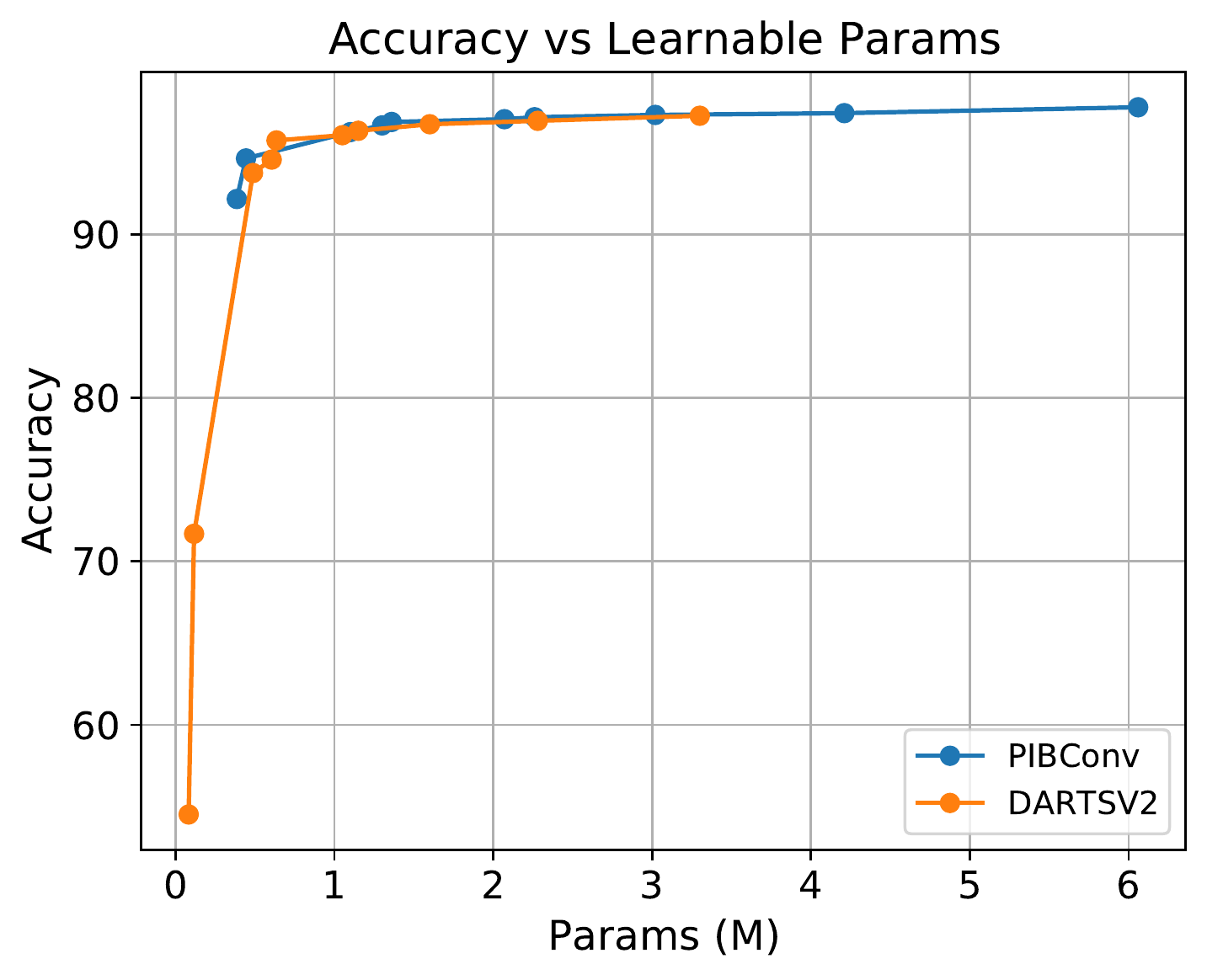}
    \hspace{0.2cm}
    \includegraphics[height=0.25\linewidth]{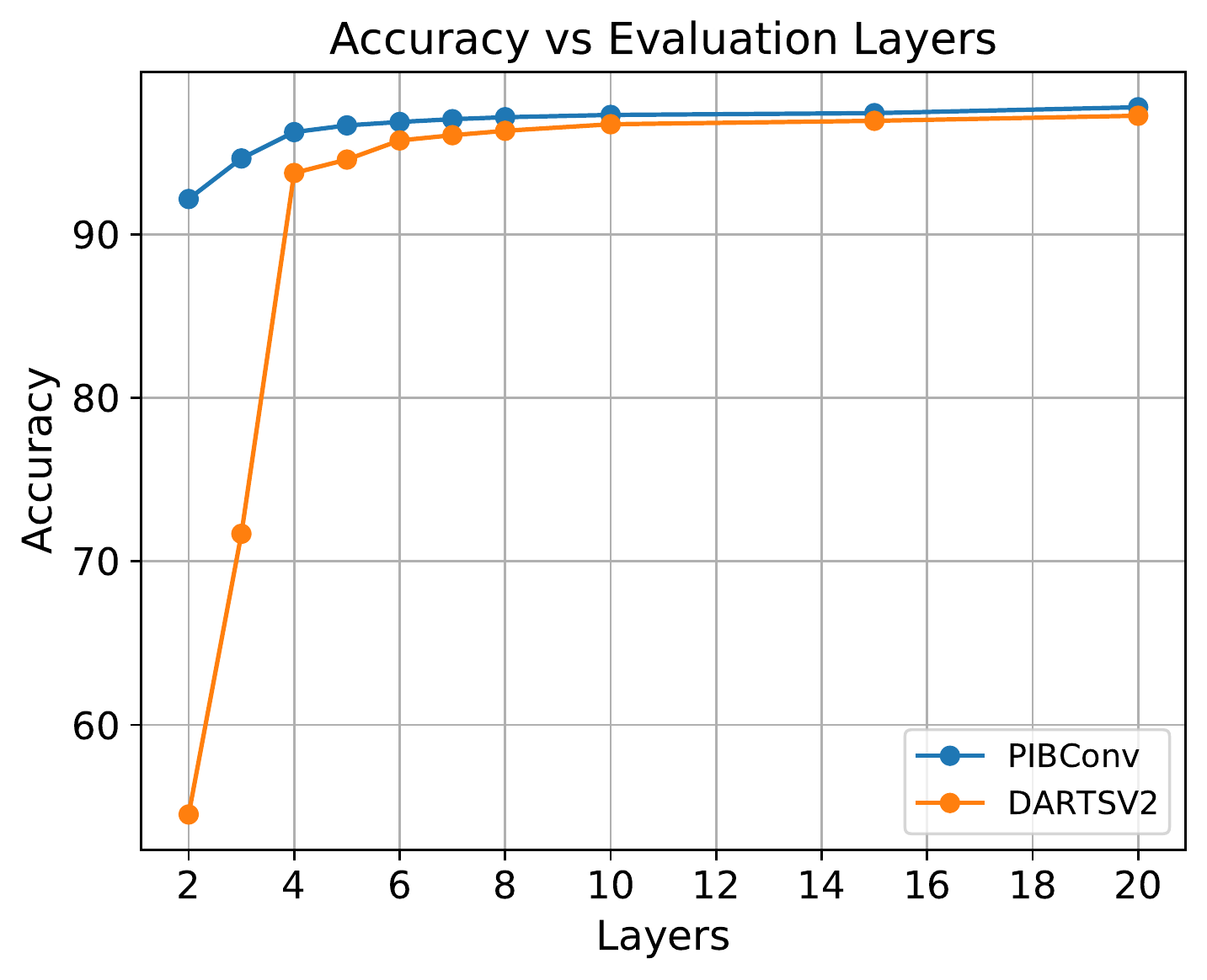}
    \caption{(a) Evolution of architecture weights if searched for 115 epochs. (b) Searched genotype using PIBConv in comparison with DARTSV2: Accuracy vs. Learnable Parameters, and (c) Accuracy vs. Evaluation Layers}
    \label{fig:arch-weights1}
\end{figure*}

\begin{figure}[htp]
    \label{fig:genotype}
    \graphicspath{ {./images/}}
    \centering
    \includegraphics[width=0.95\linewidth]{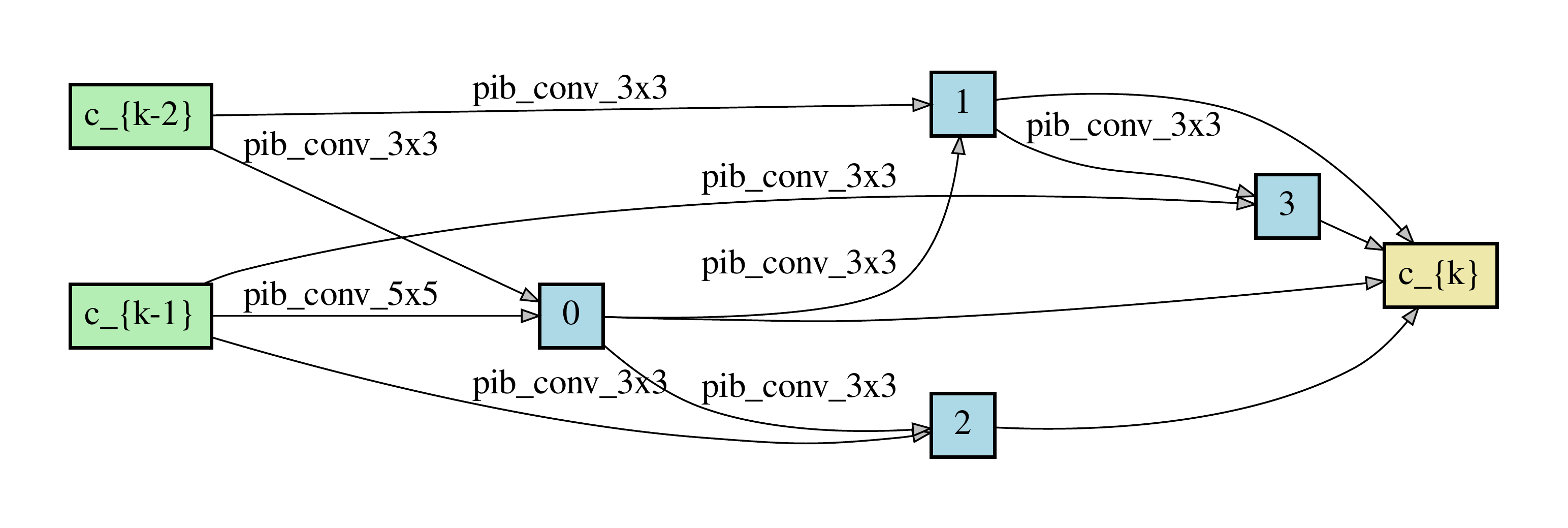}\label{fig:darts_norm}
    \includegraphics[width=0.95\linewidth]{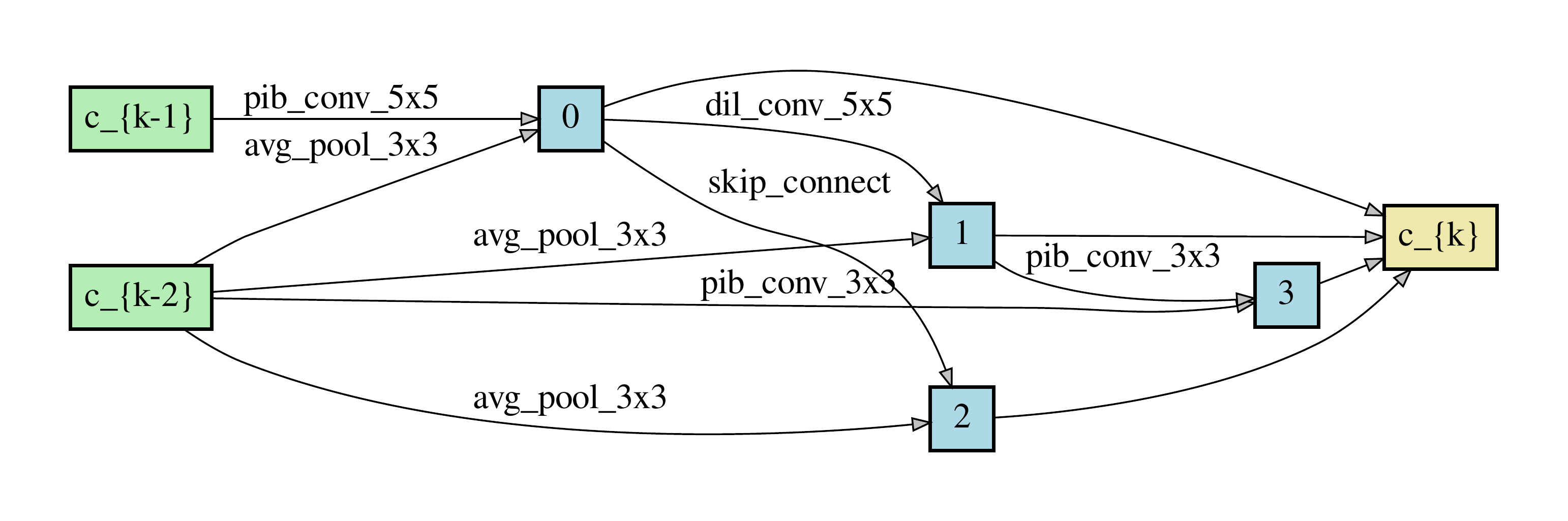}\label{fig:darts_red}
    \caption{Proposed Genotype: \textbf{(a)} Normal cell; \textbf{(b)} Reduction cell}
    \label{fig:proposed_genotype_blk}
\end{figure}

\textbf{Search Phase} Our final operation set after the incremental changes described previously is comprised of the following $10$ operations:  \textit{none}, \textit{skip\_connect}, 
\textit{pib\_conv\_3x3}, \textit{pib\_conv\_5x5},  \textit{pib\_conv\_7x7},  \textit{dialated\_conv\_3x3},  \textit{dialated\_conv\_5x5}, \textit{conv\_7x1\_1x7}, \textit{max\_pool\_3x3},  \textit{avg\_pool\_3x3}. We argue that our genotype is trained to convergence with 50 epochs and avoids a common pitfall of falling back on skip-connections in later stages of training \cite{darts_plus}. As depicted by Fig. \ref{fig:arch-weights1}, the decision boundary between the favored operation (in this case, pib\_conv\_5x5) and skip-connection, is not crossed even very late into training. After searching with the mentioned hyperparameters and final operation set, we arrive at the genotype in Fig. \ref{fig:proposed_genotype_blk}.

\begin{figure}[htp]
    \graphicspath{ {./images/}}
    \centering
    \includegraphics[width=0.175\linewidth]{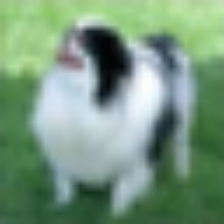}
    \includegraphics[width=0.175\linewidth]{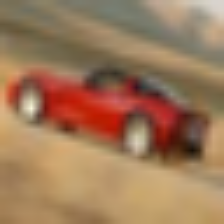}
    \includegraphics[width=0.175\linewidth]{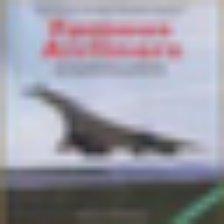}
    \includegraphics[width=0.175\linewidth]{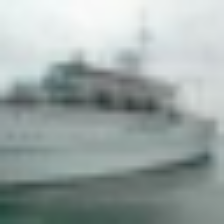}\\
    \includegraphics[width=0.175\linewidth]{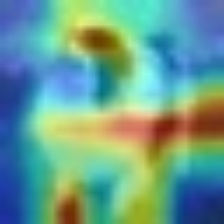}
    \includegraphics[width=0.175\linewidth]{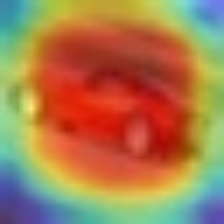}
    \includegraphics[width=0.175\linewidth]{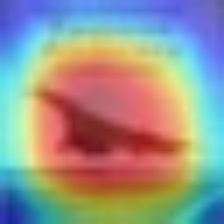}
    \includegraphics[width=0.175\linewidth]{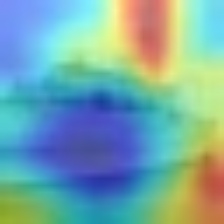}\\
    \includegraphics[width=0.175\linewidth]{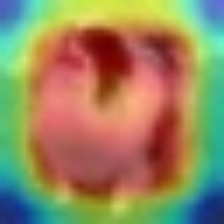}
    \includegraphics[width=0.175\linewidth]{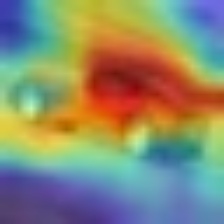}
    \includegraphics[width=0.175\linewidth]{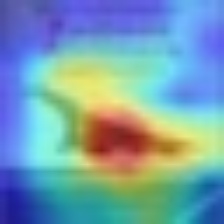}
    \includegraphics[width=0.175\linewidth]{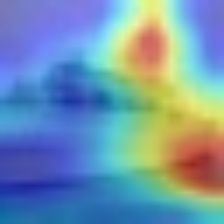}\\
    \includegraphics[width=0.175\linewidth]{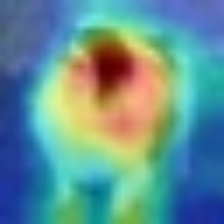}
    \includegraphics[width=0.175\linewidth]{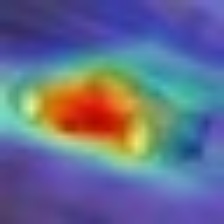}
    \includegraphics[width=0.175\linewidth]{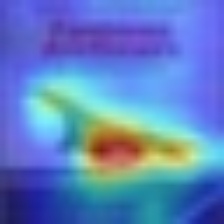}
    \includegraphics[width=0.175\linewidth]{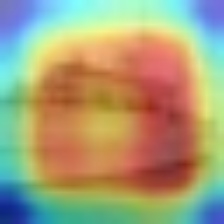}
    \caption{GradCAM: The first row shows the $32 \times 32$ input images with labels: \textit{dog, automobile, airplane, ship}; The second row shows DARTSV2 evaluated on 20 layers; Then third and fourth rows show our genotype evaluated on \{10, 20\} layers, respectively. (Note: All of the images are up-sampled to $224 \times 224$ for better readability)}
    \label{fig:GradCAM}
\end{figure}

\textbf{Evaluation Phase} We evaluate our final genotype at multiple evaluation layers to observe the effect of layer count on test accuracy and report the results in Table  \ref{tab:comparison}. We observe that the evaluation accuracy of our proposed genotype is significantly less affected by the evaluation layer count compared to DARTSV2. Specifically, at \textit{10 layers}, we achieve a higher test accuracy compared to a $20$ layer DARTSV2 network. Furthermore, at $2$ layers, our architectures exceed the DARTSV2 genotype at $3$ layers by over $20\%$, while at the same time maintaining similar GMACs. At $4$ layers, we outperform the DARTSV2 genotype at $7$ layers (to match the model size for a fair comparison) by $0.24\%$, while still maintaining lower GFLOPs. Fig.  \ref{fig:GradCAM} presents a comparison between the GradCAM \cite{gradcam} visualizations produced from the last cell of each network for DARTSV2 at $20$ layer, Our genotype at $10$ and $20$ layers. Our proposed genotype, in a $10$ cell network, can effectively capture the prominent features of the classification. The increase in the number of cascaded cells leads to the gradual collapse of the heat-map boundaries, onto the outline of the object outperforming DARTS. We argue that this supports our claim that the proposed genotype, is inherently superior to that of DARTS. 

\begin{table}[!htbp]
\centering
    \caption{Performance comparison of different genotypes on CIFAR-10 dataset: Our genotype evaluated on $10$ and $5$ layers are highlighted to be compared with DARTSV2 genotype evaluated with $20$ layers.}
    \label{tab:comparison}
    \scriptsize
    \begin{tabular}{p{0.1\linewidth}llll}
    \toprule
    \textbf{Genotype}  & \textbf{Eval. Layers} & \textbf{Test Acc. (\%)} & \textbf{Params (M)}  & \textbf{GMAC} \\
    \hline
    \midrule
    \addlinespace
    DARTSV2                             & 20    & 97.24           & 3.30  & 0.547 \\
                                        & 15    & 96.93           & 2.28  & 0.408 \\
                                        & 10    & 96.72          & 1.6   & 0.265 \\
                                        & 8     & 96.32           & 1.15  & 0.207 \\
                                        & 7     & 96.05           & 1.05  & 0.180 \\
                                        & 6     & 95.73           & 0.635  & 0.153 \\
                                        & 5     & 94.56           & 0.605 & 0.121 \\
                                        & 4     & 93.74           & 0.487  & 0.090 \\
                                        & 3     & 71.68           & 0.116  & 0.067 \\
                                        & 2     & 54.52           & 0.082  & 0.035 \\
     \addlinespace
    \hline
     \addlinespace
    PIBConv   & 20   & 97.76        & 6.06 & 0.969 \\
                                        & 15   & 97.40       & 4.21 & 0.724 \\
                                        \rowcolor{LightCyan} &\textbf{10}    & \textbf{97.29}        &\textbf{3.02} & \textbf{0.470} \\
                                        & 8    & 97.15       & 2.26 & 0.369 \\
                                        & 7    & 97.03        & 2.07 & 0.320 \\
                                        & 6    & 96.86        & 1.36 & 0.275 \\
                                       \rowcolor{LightCyan} &\textbf{5}    & \textbf{96.65}        & \textbf{1.30} & \textbf{0.218} \\
                                        & 4    & 96.24        & 1.10 & 0.166 \\
                                        & 3    & 94.63        & 0.443 & 0.123 \\
                                        & 2    & 92.15       & 0.385 & 0.067 \\
    \addlinespace
    \hline
    \bottomrule
    \end{tabular}
\end{table}

\begin{figure}[htp]
    \graphicspath{ {./images/}}
    \centering
    \includegraphics[scale=0.5]{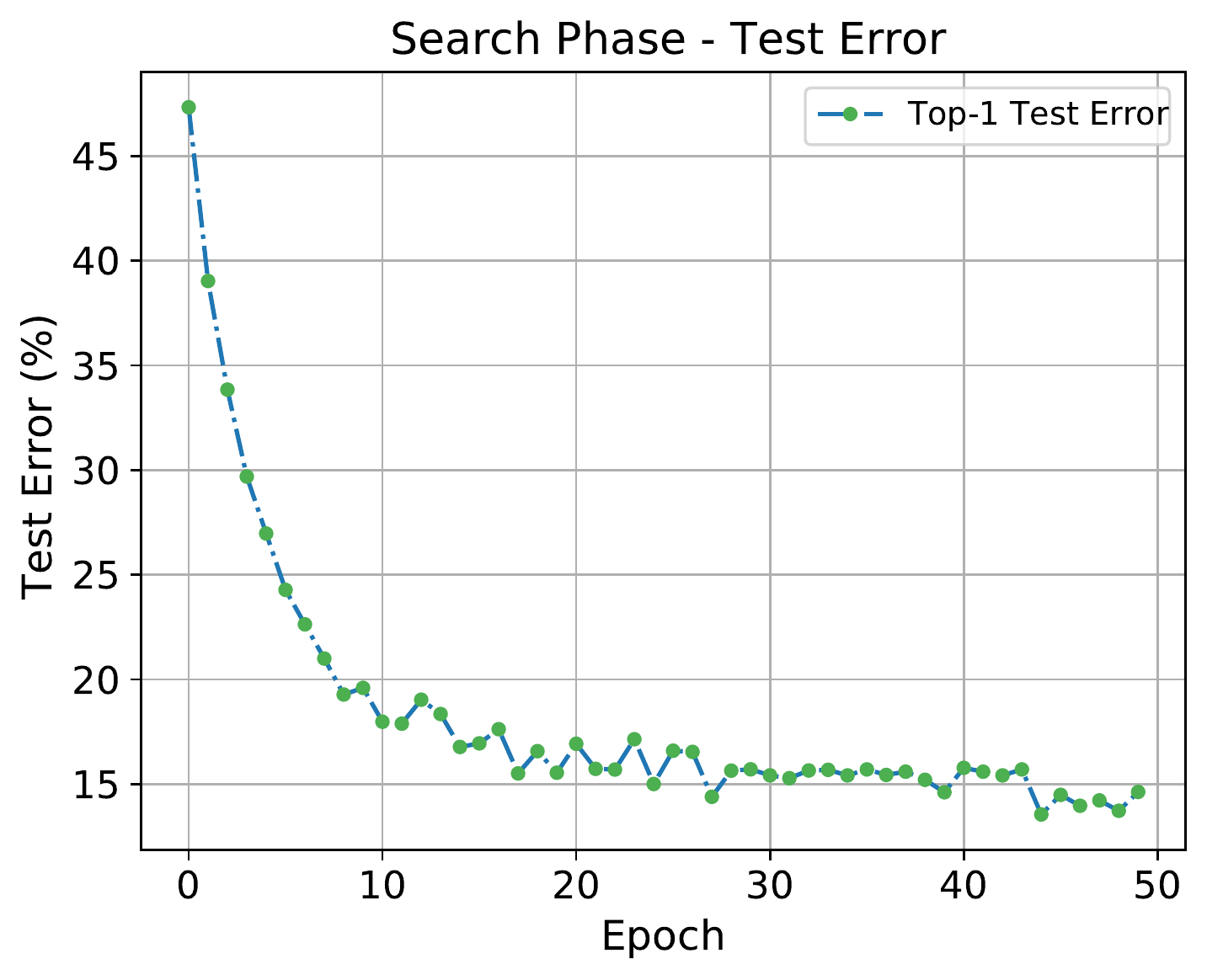}
    \caption{Search phase of our proposed genotype: Top $1\%$ Test Error vs Epoch}
    \label{fig:arch-weights}
\end{figure}

\section{Conclusion}
In this work, we attempt to revise the DARTS search space. We incrementally augment the convolution operation with micro-changes inspired by ConvNeXt and propose the Pseudo-Inverted Bottleneck block to reduce the number of parameters used in the vanilla Inverted Bottleneck. Our proposed genotype's performance is much less sensitive to evaluation layer count compared to that of DARTSV2. It achieves a higher accuracy at a lower GMAC/ parameter count with $10$ evaluation layers compared to DARTSV2 evaluated at $20$ layers. Furthermore, we perform a GradCAM visualization on our genotype and compare it with that of DARTSV2.

Our network's high performance at lower layer counts, correspondingly with low GMACs and parameter count, makes it an attractive choice for (a) image processing applications such as sharpening and blurring, as shallow networks suit these applications best; and (b) designing lightweight network design framework for efficient representation learning on edge devices. Consequently, a potential avenue for future work would be to explore the applications of our genotype/ Pseudo-Inverted Bottleneck block, in both low-level and high-level vision processing tasks. 

It is worth noting that our aim in this paper was not to combat the SOTA methods related to DARTS (which deems to be the limitation of our work here); but shedding light on the granularity of search space which is commonly shared across many DARTS variants in the literature. We hope our work initiates new ideas to investigate optimum search space designs in DARTS framework to build more robust and generalized models for representational learning problems.\FloatBarrier

\bibliography{references.bib}
\end{document}